\documentclass[10pt,twocolumn,letterpaper]{article}

\usepackage{cvpr}
\usepackage{times}
\usepackage{epsfig}
\usepackage{graphicx}
\usepackage{amsmath}
\usepackage{amssymb}
\usepackage{epstopdf}

\graphicspath{{./}}

\usepackage[pagebackref=true,breaklinks=true,letterpaper=true,colorlinks,bookmarks=false]{hyperref}

\cvprfinalcopy 

\def\cvprPaperID{1146} 

\ifcvprfinal\fi
\begin{document}

\title{Spatially Coherent Random Forests}

\author{Tal Remez\qquad Shai Avidan\\Tel-Aviv University, Department of Electrical Engineering}


\maketitle

\begin{abstract}
Spatially Coherent Random Forest (SCRF) extends Random Forest to create spatially coherent labeling. Each split function in SCRF is evaluated based on a traditional information gain measure that is regularized by a spatial coherency term. This way, SCRF is encouraged to choose split functions that cluster pixels both in appearance space and in image space. In particular, we use SCRF to detect contours in images, where contours are taken to be the boundaries between different regions. Each tree in the forest produces a segmentation of the image plane and the boundaries of the segmentations of all trees are aggregated to produce a final hierarchical contour map. We show that this modification improves the performance of regular Random Forest by about $10\%$ on the standard Berkeley Segmentation Datasets. We believe that SCRF can be used in other settings as well.
\end{abstract}

\section{Introduction}
Random Forest (RF) is a popular method to solve multi-label inference problems such as regression, classification and clustering. Most notably, it has shown remarkable results on pixel labeling problems where an image is to be segmented into a predefined number of categories.

RF is easy to implement, scales well with the amount of data available and is efficient in test time. The downside of RF methods is that they work on each pixel independently. As a result, pixel labeling must be smoothed, usually in a post-processing step.

We are interested in applying RF to the problem of contour detection in natural images, as this is a surrogate to the image segmentation problem. Alas, we need to distinguish between clustering and segmentation. In clustering each pixel (or patch) must be assigned to a cluster but the spatial position of the pixel is irrelevant. This can result in a highly incoherent layout in the image plane. Segmentation, on the other hand, seeks to cluster pixels both in appearance space and the image plane.

\begin{figure}
    \centering 
        \begin{tabular}{c}
        	 \includegraphics[width = 0.4\textwidth]{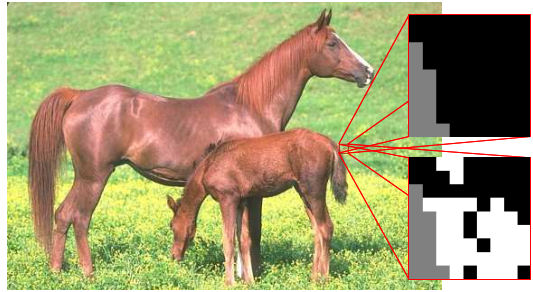} \\
        \end{tabular}   \\ 
    \caption{\small \textbf{Spatially Coherent Random Forest (SCRF)} The role of the spatial coherency term. Given the input image and a particular node, consider two split functions (top and bottom zoom-ins). Each split function generates a map where gray pixels are pixels that did not reach the current node, white and black pixels are pixels that are assigned to the left and right children of the current node, respectively. The top split function generated a much smoother segmentation for this particular $9 \times 9$ patch.}
\label{fig_spatial_term}
\end{figure}

We propose Spatially Coherent Random Forest (SCRF) as a way to add spatial reasoning to RF. SCRF defines a new evaluation function for each node in the tree. The new function measures a regularized expression that combines information gain (as is done in traditional RF) with a spatial coherency term. The spatial coherency term counts the number of pixels that are assigned to the same cluster within a patch. The resulting tree produces a spatially coherent segmentation of the image plane. Figure~\ref{fig_spatial_term} demonstrates the idea graphically.


Each tree in SCRF segments the image and we use the boundaries between segments to produce a contour map, per tree. The contour maps of all trees in the forest are then combined to give the final contour map of the image. We refine this basic algorithm by creating boundary maps for each level of the tree. This leads to a hierarchical boundary map.

SCRF requires a very subtle change to the way RF are used. This change does not hurt the good properties that led to the popularity of RF, yet it improves results considerably. We demonstrate the effectiveness of the proposed change on a couple of synthetic experiments and show that SCRF is about $10\%$ better than RF on the standard BSD300 and BSD500 datasets.

\section{Background}

Our work deals with the application of Random Forests to the problem of contour detection in natural images. For a comprehensive overview of RF see \cite{criminisi2013decision}.


RF are scalable, versatile and efficient. On the downside, they treat each pixel independently, which prevents them from producing spatially coherent labeling.
To overcome this problem a post processing is usually done on the output of the forest by a Markov or Conditional random field (MRF/CRF) \cite{blake2011markov} \cite{nowozin2011decision}. For instance, in \cite{shotton2009textonboost} the results of the first forest are post-processed by a CRF. Alternatively, the computationally heavy CRF smoothing can be replaced by a second forest \cite{shotton2008semantic}.

Kontschieder {\em et al.} \cite{kontschieder2011structured} also look at the spatial distribution of labels in a patch around each pixel and modify the information gain accordingly. Specifically, they enforce spatial structure, during training, by using a labeling for each pixel that is drawn uniformly from the patch around each pixel once per node.  The spatial consistency of the segmentation was later improved by encoding variable dependencies directly in the feature space the forests operate on \cite{KontschiederKSC13}.

The approaches mentioned above lead to impressive results but focused mainly on the supervised setting of semantic segmentation, where a pixel is to be assigned to one of a predefined list of categories.

Forests have also been used in interactive segmentation, as in \cite{santner2009interactive}, where a segmentation is generated using an iterative process involving forest pixel labeling followed by a weighted Total Variation regularization using a massively parallel GPU implementation.

Video segmentation enjoyed the benefits of random forest as well. Perbet {\em et al.} \cite{perbet2009random} presented a clustering algorithm based on random forests and Markov clustering, shown to work well for video segmentation applications. Yin {\em et al.} \cite{yin2007tree} segmented video into foreground and background layers based on a combination of a random forest and a CRF. Recently, Criminisi {\em et al.} \cite{criminisi2013decision} proposed to use random forest for density estimation as a proxy to unsupervised tasks.

As far as contour detection goes, most recent approaches take into account texture and color features \cite{martin2004learning}, \cite{mairal2008discriminative}. Learning techniques have also been adapted to contour detection. For example, the Boosted Edge Learning \cite{dollar2006supervised} algorithm makes use of a supervised learning phase in which it attempts to learn an edge classifier in the form of a probabilistic boosting tree from thousands of simple features computed on image patches. Multi scale approaches attempted to improve segmentation \cite{ren2008multi} by improving robustness to the range of scales in which objects may appear in natural images.

A different approach to contour detection is based on global cues rather than local ones. Early algorithms {\cite{parent1989trace},\cite{elder1996computing}} link together high-gradient edge fragments in order to identify extended, smooth contours. More recent approaches such as \cite{ren2005scale} use a CRF to enforce curvilinear continuity of contours. Felzenszwalb and McAllester \cite{felzenszwalb2006min} suggested a method for finding salient curves in an image using a min-cover framework.

Curve detection can be seen as a by product of image segmentation. In particular, an image is reduced to a graph representation and there is a wealth of algorithms for the optimal partitioning of a graph, e.g. Normalized Cuts \cite{ShiM00} or, more recently,  multi-cuts \cite{andres2011probabilistic} where image partitioning is represented as an edge labeling problem and integer linear programming is used  to iteratively solve a set of constraints. Alternatively, one can take a bottom up approach, as suggested by Felzenszwalb and Huttenlocher \cite{felzenszwalb2004efficient} who proposed a predicate for measuring the evidence for a boundary between two regions and a greedy algorithm based on it. They showed that although the algorithm makes greedy decisions it produces segmentations that satisfy global properties.

Finally, Arbelaez {\em et al.} \cite{arbelaez2011contour} tackle both contour detection and image segmentation. Contour detection is based of combining multiple local cues into a global spectral clustering algorithm. Their segmentation algorithm transforms the estimated contour map into a hierarchical region tree.

\subsection{Random Forests}

Random Decision Forests are designed to deal with labeled data. A decision forest is an ensemble of $T$ decision trees. A node $n$ in a tree holds an estimated class distribution $P(c|n)$ of classes $c$. A decision tree works by recursively branching left or right down the tree according to a learned binary split function of the feature vector, until a leaf node $l$ is reached. The forest classifies a pixel using the average class distribution of all the leaves that pixel reached $L = (l_1,...,l_T)$:
\begin{equation} \label{equ_forest_avg}
P(c|L) = \frac{1}{T} \sum_{t=1}^{T}{P(c|l_t)}
\end{equation}
Let $S_n$ denote the set of all pixels reaching node $n$. Then $S_n$ is split, by split function $f$, into two mutually exclusive subsets, the right subset $S_{n,f}^R$ and the left subset $S_{n,f}^L$. In our case, the split function operates on the value of a pixel at location $p$ and channel $b$:
\begin{equation} \label{equ_forest_avg}
S_{n,f}^L = \{p \in S_n | f(p,b)<t\}
\end{equation}
\begin{equation} \label{equ_forest_avg}
S_{n,f}^R = S_n \setminus S_{n,f}^L
\end{equation}
At each non-leaf node, many pairs of candidate split functions and their thresholds are generated randomly. The pair that maximizes the expected gain in information is chosen:
\begin{equation} \label{equ_infogain}
I_n(f) = H(S_{n}) - \sum_{a \in \{L,R\}}{\frac{|S_{n,f}^a|}{|S_{n}|}H(S_{n,f}^a)}
\end{equation}
We denote by $H(S)$ Shannon's entropy of the classes in the set of examples $S$. Training is continued until a maximal depth $D$ is reached or until no information gain is possible. The class distribution $P(c|n)$ of a node is estimated empirically with histograms of class labels $c_i$ of the training examples that reach node $n$.

When labeled data is not available, one can use Density estimation forests \cite{criminisi2011decision} to estimate the latent probability density function from a set of given data points. And because we do not have labeled data we use multi-variate Gaussian distributions at the nodes, instead. The differential entropy of a multi-variate Gaussian can be shown to be:
\begin{equation} \label{equ_info_gaussian}
H(S) = \frac{1}{2}\log{((2\pi e)^m |\Lambda(S)|)}
\end{equation}
(where $\Lambda$ is the $m\times m$ covariance matrix).
Consequently, the information gain \ref{equ_infogain} in the unsupervised case becomes:
\begin{equation} \label{equ_infogain_gaussian}
I_{n}(f) =log{|\Lambda(S_{n,f})|} - \sum_{a \in \{L,R\}}{\frac{|S_{n,f}^a|}{|S_{n,f}|}log{|\Lambda(S_{n,f}^a)|}}
\end{equation}
The size of the covariance matrix $\Lambda$ is $m\times m$ where $m$ in the number of image channels. This could become computationally expensive if the number of channels is large. In addition, as in any multidimensional clustering problem many of the dimensions are mostly irrelevant. Therefore when calculating the information gain for each split function only the image channel $b$ used by that split function is taken into account in \ref{equ_info_gaussian}. The other channels do not affect the information gain introduced by the split function. This reduces computational cost and improves accuracy.

\section{Spatially Coherent Random Forest}

Spatially Coherent Random Forests (SCRF's) are RF's that produce spatially coherent clusters. They differ from regular RF's in the way each node evaluates its split function. While in regular RF's a split function is chosen solely based on an information gain, we propose to add a regularization term that represents the spatial coherency of the resulting split in the image plane. Weighting the spatial coherency term with the information gain affects the split functions that are chosen at each node. Increasing the weight will bias the selection toward split functions that result in smoother splits in the image plane and hence smoother pixel labelling or image segmentations.

\begin{figure*} [t]
    \centering 
        \begin{tabular}{ c  c  c}
            \includegraphics[width = 0.215\textwidth]{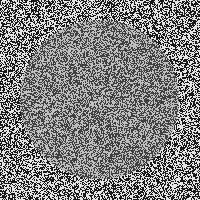} &
            \includegraphics[width = 0.23\textwidth]{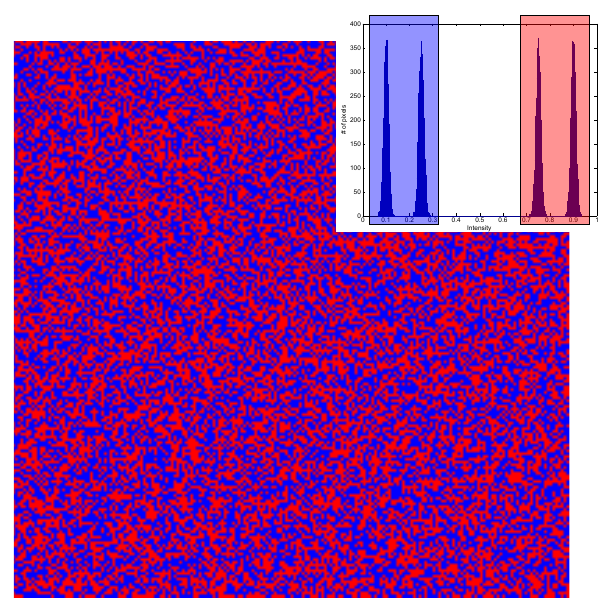} &
            \includegraphics[width = 0.23\textwidth]{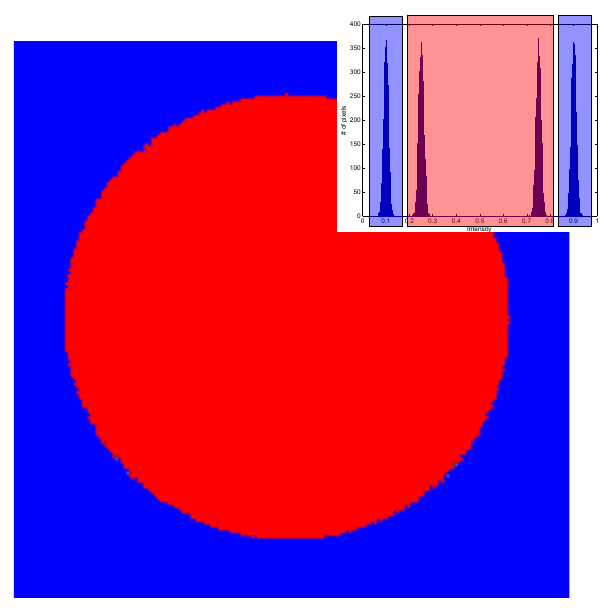} \\
            (a) Input & (b) Regular Random Tree & (c) Spatially Coherent Random Tree\\

        \end{tabular}   \\ 
    \caption{\small \textbf{Synthetic segmentation example}. (a) input image (see text for details), (b) segmentation result of a regular random tree (segmentation of histogram shown in inset) and (c) segmentation of our spatially coherent random tree. (segmentation of histogram shown in inset).}
    \label{fig_toy_1}
\end{figure*}

\subsection{The Spatial Coherency Term}
Intuitively, while segmenting or labeling an image using a RF we would expect pixels that belong to the same segment or object to follow similar paths down the forest's trees. In addition, because natural images are usually piecewise smooth, we expect that the splitting of the image, while moving down a tree, would reflect this piecewise smoothness via spatially smooth splits. Thus, the objective of our spatial coherency term is to quantify how similar the faith of a pixel is in comparison to its neighboring pixels.

Given a pixel $p$ reaching node $n$ we look at a $r\times r$ window, $w_p$, centered around it. We can divide the set of pixels in $w_p$ into 3 mutually exclusive sets. Pixels that reach node $n$ and are assigned to the left child, pixels that reach node $n$ and are assigned to the right child, and pixels that did not reach node $n$ at all.

Given node $n$, we wish to evaluate a split function $f$ that can assign each pixel $p$ that reached $n$ into one of two clusters ${\{L,R\}}$. We take the spatial coherency term, termed $C_n(f,p)$, to be the fraction of pixels that were assigned to the same label as pixel $p$, relative to the size of $w_p$.
\begin{equation} \label{equ_spatial_pixel_area}
C_n(f,p) = \frac{|\{q| (q \in S_n^{f,a}) \wedge (p \in S_n^{f,a}),~~ a \in \{L,R\}\}|}{r^2}
\end{equation}
That is, we require pixels $p$ and $q$ to be assigned to the same child. Also note that we have $C_n(p)\in[0,1]$. Adding $C_n(f,p)$ to the information gain \ref{equ_infogain_gaussian} we get the overall score of split function $f$ at node $n$:
\begin{equation} \label{equ_total_grade}
G_{n}(f) = I_{n}(f) + \lambda \cdot \sum_{p \in n} C_{n}(f,p)
\end{equation}

\subsection{SCRF for Image Segmentation} \label{par_forest_seg}

To illustrate the power of SCRF we generated a few synthetic examples and segmented them using our spatially coherent segmentation forest. Figure~\ref{fig_toy_1} gives one such example of a forest with a single tree. It contains two regions, each of which was generated using a Gaussian Mixture Model (GMM) with two equally probable Gaussians with identical standard deviations. The two Gaussians of the central circle are separated by the two Gaussians of the background. Therefore, any attempt to segment the image into two regions solely based on the image intensity will result in a meaningless segmentation. Trying to segment this image using a regular random forest fails as well due to the greedy nature of the split function selection. According to equation \ref{equ_infogain_gaussian}, a greedy information based selection will divide the image into two regions such that each of these regions has the lowest intensity variance, while introducing the spatial coherence term as in \ref{equ_total_grade} will result in much smoother results as can be seen.



Each tree in the SCRF defines a segmentation of the image, where each segment is defined as all pixels reaching a specific leaf of the tree. Therefore, we have as many segmentations of the image as we have trees in the forest. There is a large body of work on how to combine multiple segmentations into a single, coherent result. We are not pursuing this direction here. Instead, we focus on the problem of contour detection.

\subsection{SCRF for Contour Estimation} \label{par_forest_contour}

Each tree segments the image and we would like to merge all $T$ trees in the forest to give one coherent contour map of the image. The first step is to estimate the contour map of a single tree segmentation. Then we can generate a contour map of the forest by averaging the maps of the forest's $T$  trees. The averaging of the contour maps adds robustness to the estimation and prevents the random nature of each tree from heavily affecting the global contour map.

While trying to generate a contour map of an image using a single tree all the information that can be gained from that tree should be taken into account. For instance, using only the segmentation defined by the leafs of the tree gives a flat contour map that does not emphasize the importance of different contours. Our underlying assumption is that we can learn something about the image contour from each split in the tree and by the depth of that split.

Given node $n$, the data $S_n$ associated with it and a split function $f$, let pixel $p$ belong to $S^a_{n,f}$ for some $a \in \{L,R\}$. Then the edge map $E_n$ associated with node $n$ is:
\begin{equation}
E_n(p) = \left\{ \begin{array}{cc}
				1 & \exists q \in {\cal N}(p)~\mbox{s.t.}~q \in S_n \setminus S^a_{n,f}\\
				0 & \mbox{otherwise}\\
				\end{array} \right.
\end{equation}
where ${\cal N}(p)$ denotes the neighborhood of $p$ in the image plane. In words, a pixel $p$ is an edge pixel of node $n$, if it has neighbor pixel $q$ (in the image plane) that arrived at node $n$ and was assigned to a different cluster. The edge map of a particular level $d$ of the tree is given by:
\begin{equation}
E_d(p) = max \{E_n(p) \}_{depth(n)=d}
\end{equation}
And the overall contour map is:
\begin{equation}
E(p) = \sum_{d=0}^D \frac{D-d}{D} E_d(p)
\end{equation}
The term $\frac{D-d}{D}$ gives higher weight for tree nodes that are closer to the root. We use the $max$ operator to evaluate the edge map of a particular level in the tree to avoid double counting of pixels. But when we sum across levels, we do want to emphasize contours that separate regions at multiple scales.

An alternative to the method we proposed for contour detection is to aggregate all tree segmentations into a single forest segmentation and take the contours to be the boundaries between the regions of this final segmentation. In particular, we compare our method to spectral clustering with a predefined number of clusters $K$. For each pixel $p$, we randomly sample a total of $N$ pixels from the leafs that $p$ reached in any of the trees of the forest. The affinity between a pair of pixels $p_1$ and $p_2$, is based on their paths in the trees. In particular, we take it to be a function of their last common ancestor (LCA) $l_t(p_1,p_2)$ weighted by its depth $d(l_t(p_1,p_2))$, in each tree. LCA is taken to be the last node in a tree that both pixels shared. Formally, the affinity is given by:
\begin{equation} \label{equ_affinity}
\begin{array}{ccc}
A(p_1,p_2) & = & \frac{1}{T}\sum_{t=1}^{T}{(1 - 2^{-d(l_t(p_1,p_2))}) }\\
                & = & 1 - \frac{1}{T}\sum_{t=1}^{T}{ 2^{-d(l_t(p_1,p_2))} }\\
\end{array}
\end{equation}
If two pixels were split at the root of all trees in the forest then $A(p_1,p_2)=0$ and if they reach the same leaf nodes in all trees then $A(p_1,p_2)\rightarrow 1$.

\begin{figure*} [t]
    \centering 
        \begin{tabular}{ c  c c c c }
            \includegraphics[width = 0.18\textwidth]{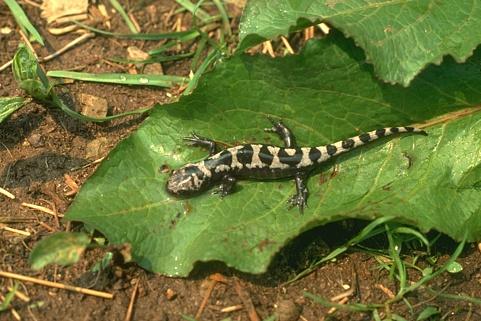} &
            \includegraphics[width = 0.18\textwidth]{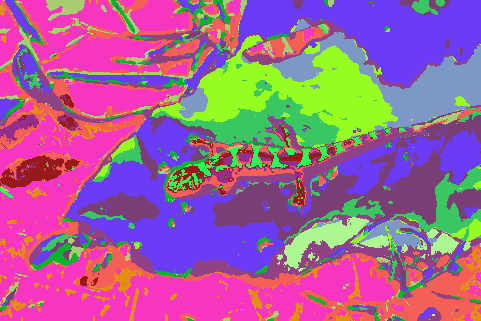} &
            \includegraphics[width = 0.18\textwidth]{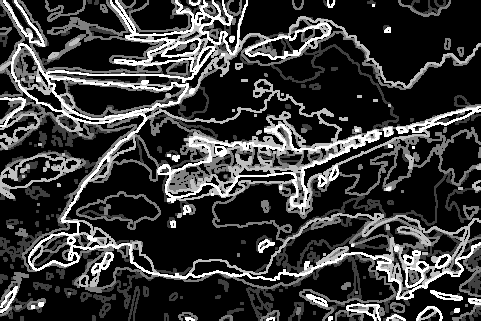} &
            \includegraphics[width = 0.18\textwidth]{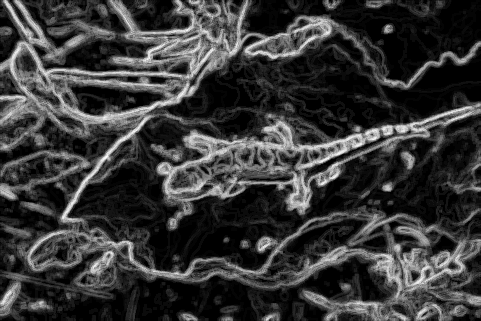} &
		 \includegraphics[width = 0.18\textwidth]{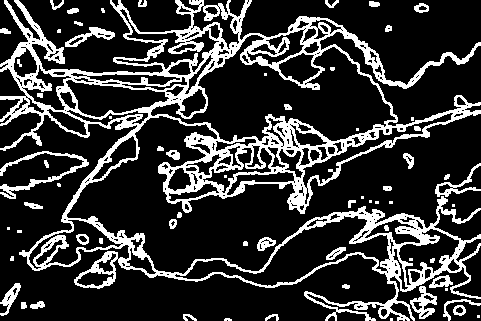} \\
            (a) Image & (b) Tree Segmentation & (c) Tree Contour & (d) Forest Contour & (e) Spectral Clustering \\
         \end{tabular}   \\ 
    \caption{\small \textbf{Contour Estimation}. (a) input image. (b) segmentation defined by the leafs of one tree. (c) contour estimation of this tree. (d) contour estimation of the entire forest. We can see how averaging contour maps of a few trees removes wrong edges and emphasizes correct ones. (e) edges after running Spectral Clustering. As can be seen, (d) is comparable to (e) but avoids the computational expensive spectral clustering step.}
    \label{fig_contour_1}
\end{figure*}

Figure \ref{fig_contour_1} compares our SCRF contour detection algorithm to the contours produced after spectral clustering. It can be seen that SCRF with forest contour estimation produce superior results to a regular density estimation forest followed by spectral clustering. In addition our method does not require spectral clustering as a computationally expensive post-processing.

\section{Experiments}

We have conducted a number of experiments to evaluate SCRF.  In all experiments we have used split functions similar to the ones used in STF \cite{shotton2008semantic}. These split functions act on small image patches and test simple relations between pixel values in this patch, these tests are:

\begin{itemize}

\item  $f(p,b)$ - The value of pixel $p$ in channel $b$.
\item  $|f(p,b)-c|$ - The distance of pixel $p$ in channel $b$ from a bias value $c$.
\item  $f(p_1,b) + f(p_2,b)$ - The sum of two pixel values from channel $b$.
\item  $f(p_1,b) - f(p_2,b)$ - The difference between two pixel values from channel $b$.
\item  $|f(p_1,b) - f(p_2,b)|$ - The absolute difference between two pixel values from channel $b$.

\end{itemize}

We sample many random split functions, and each of them selects a random image channel, to ensure a proper cover of the space of split functions. The combination of a split function and an image channel that yields the maximal information gain are chosen at each node.

\begin{figure} [t]
    \centering 
        \begin{tabular}{ c  c  }
            \includegraphics[width = 0.23\textwidth]{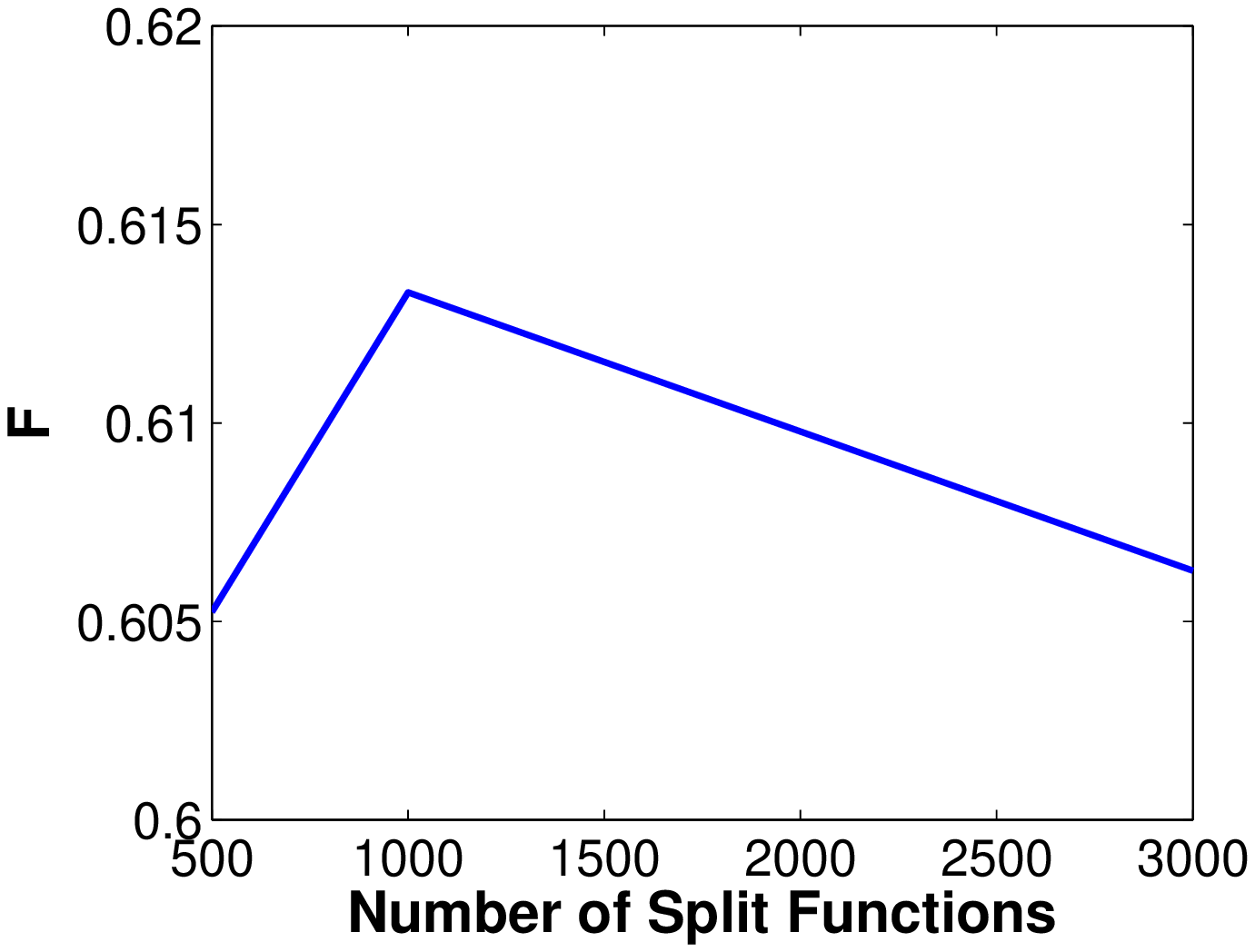} &
            \includegraphics[width = 0.23\textwidth]{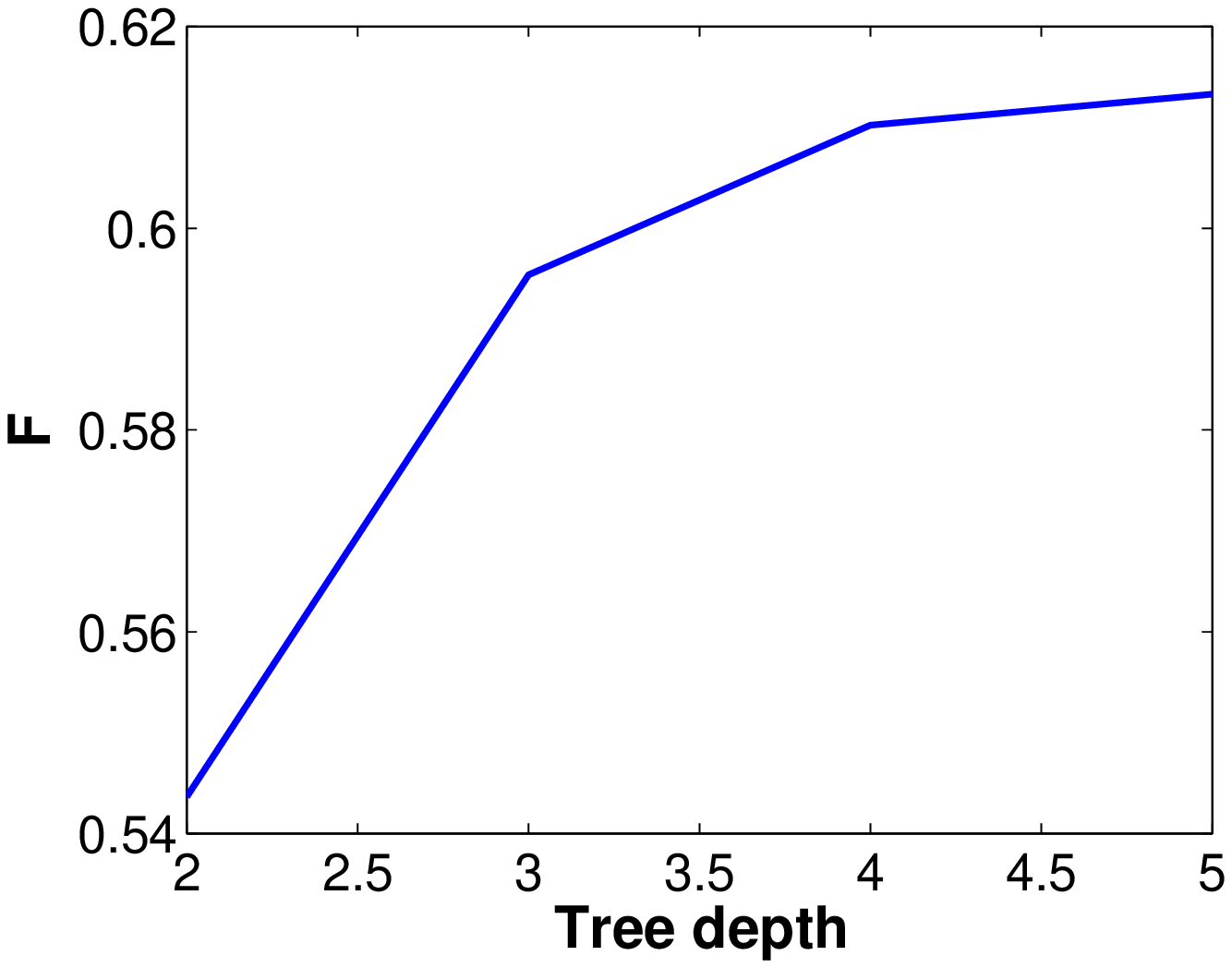} \\
            \includegraphics[width = 0.23\textwidth]{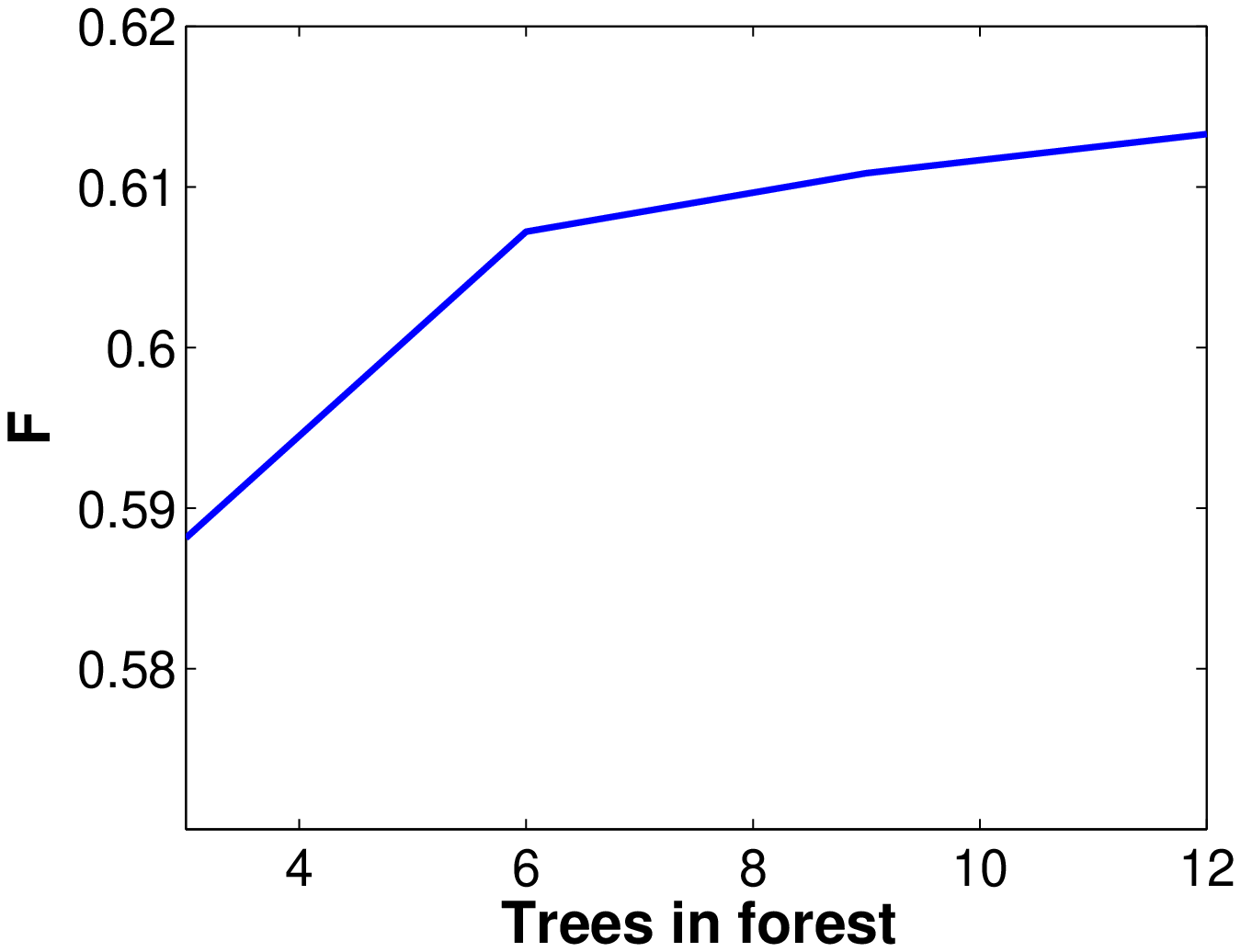} &
            \includegraphics[width = 0.23\textwidth]{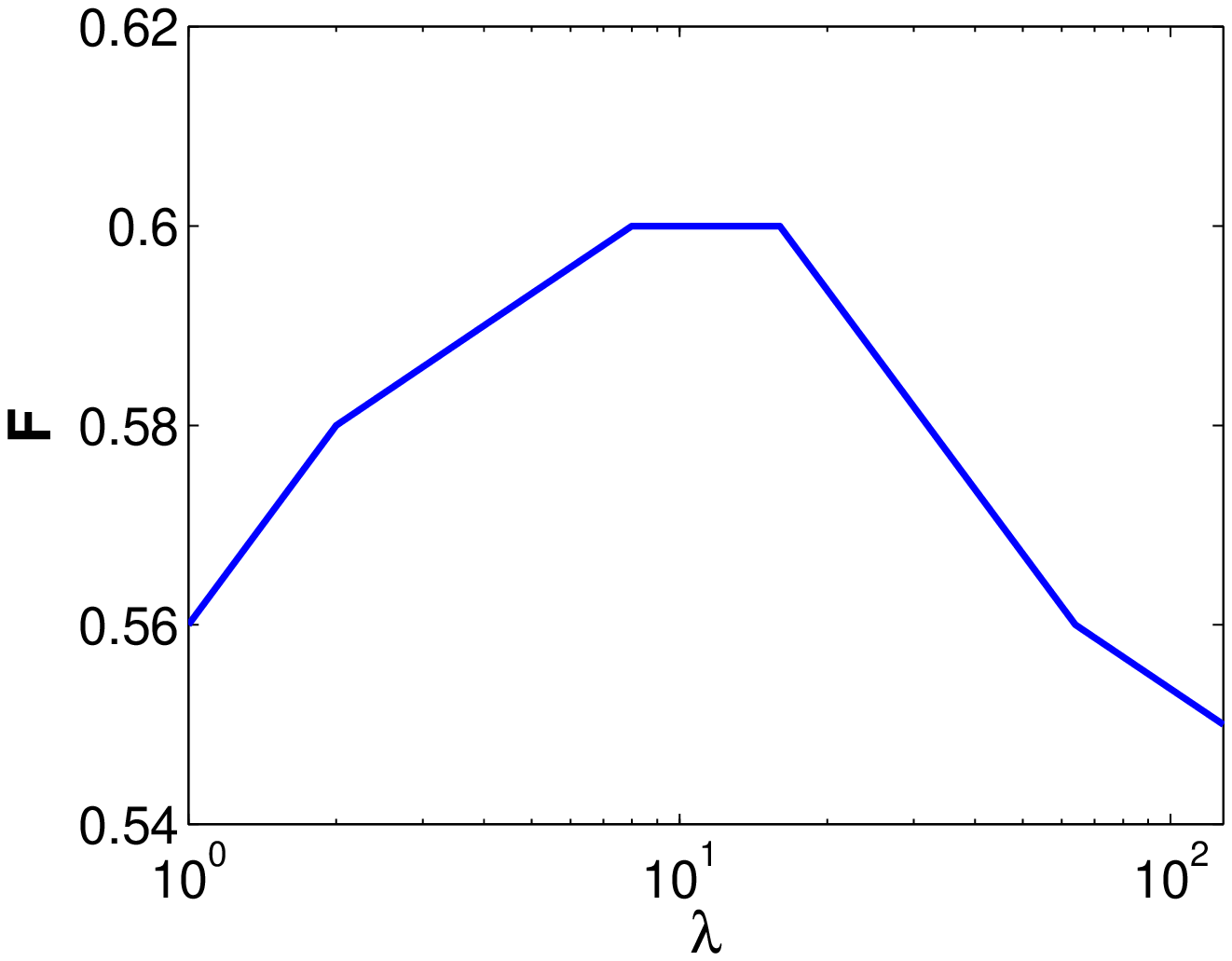} \\
         \end{tabular}   \\ 
    \caption{\small \textbf{Parameter optimization}. Presented are graphs showing the effect each parameter of the forest has on the F measure. Default parameters, except for the one being optimized, were number of trees $T=12$, depth of tree $D=5$, spatial coherency term $\lambda=8$ and 1000 split functions per node.}
    \label{fig_opt}
\end{figure}

First, we conducted a number of experiments to fine tune the parameters. Figure \ref{fig_opt} shows the effect of maximal tree depth, number of split functions at each node, the number of trees in the forest, and the weight $\lambda$ of the spatial coherency term on the F measure.

As a result of this parameter optimization we fixed the parameters to be 12 trees per forest. The maximum depth of each tree was set to 5. Split functions use patches of size $5\times 5$, and the spatial coherency term uses patches of size $9\times 9$. The minimal number of pixels per leaf was set to 1000 and at each node 1000 random split functions were evaluated. The spatial coherency parameter that gave the best results on the training images was $\lambda=8$. When segmenting an image we generate $\lambda$ randomly at each split node to prevent over-fitting, where $\lambda$ is generated from a Gaussian distribution $\lambda\sim N(8,4)$.

\begin{figure*} [t]
    \centering 
        \begin{tabular}{ c  c  c  c  c  c }
            \includegraphics[width = 0.15\textwidth]{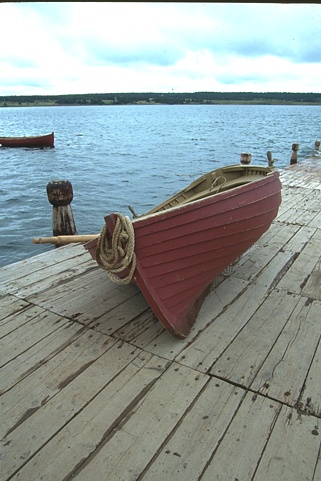} &
            \includegraphics[width = 0.15\textwidth]{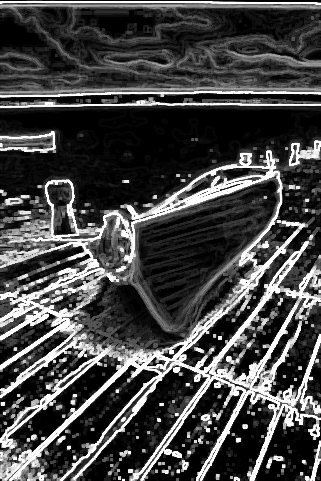} &
            \includegraphics[width = 0.15\textwidth]{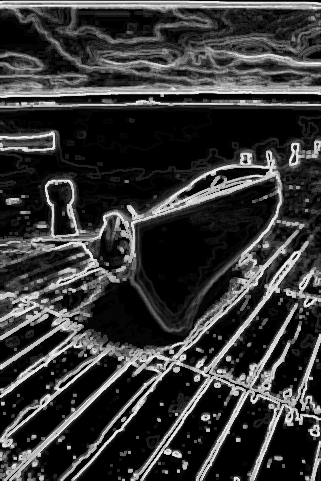} &
            \includegraphics[width = 0.15\textwidth]{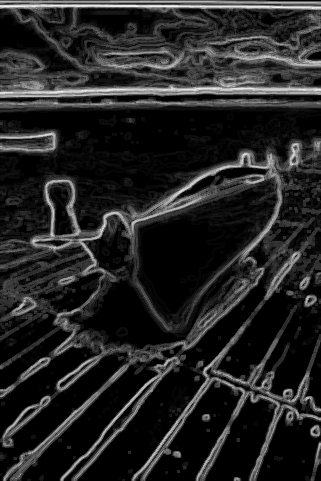} &
            \includegraphics[width = 0.15\textwidth]{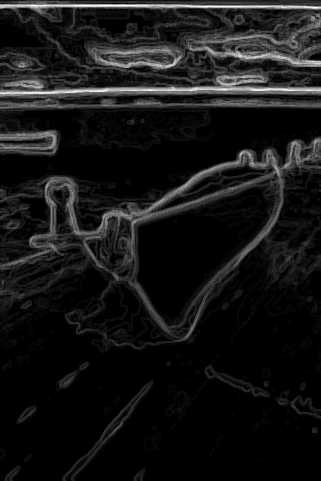} &
            \includegraphics[width = 0.15\textwidth]{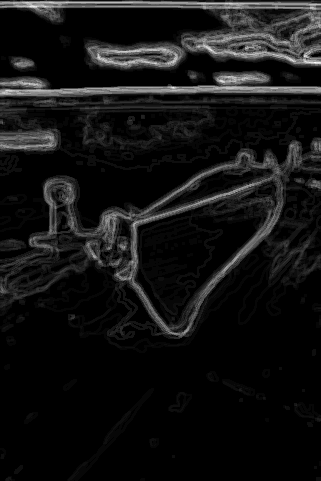} \\

            \includegraphics[width = 0.15\textwidth]{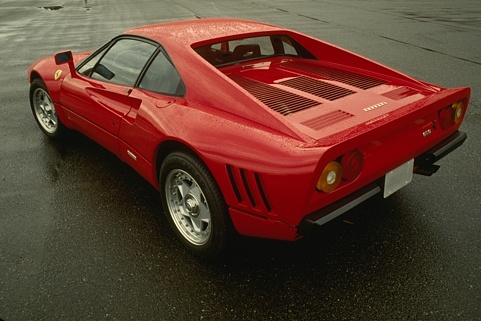} &
            \includegraphics[width = 0.15\textwidth]{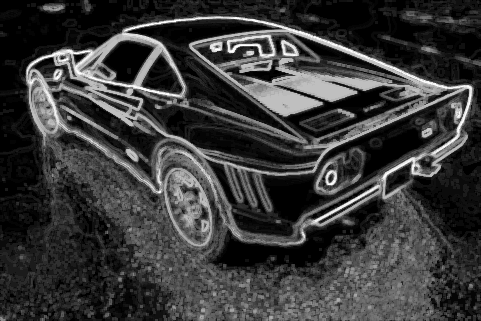} &
            \includegraphics[width = 0.15\textwidth]{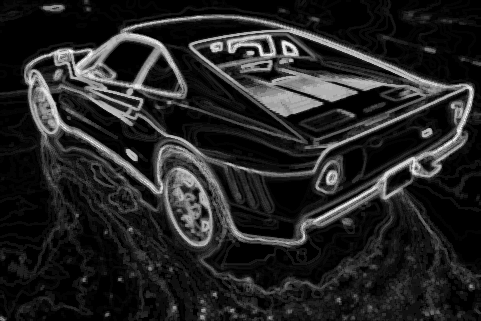} &
            \includegraphics[width = 0.15\textwidth]{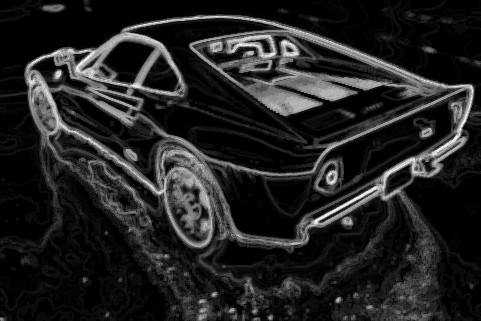} &
            \includegraphics[width = 0.15\textwidth]{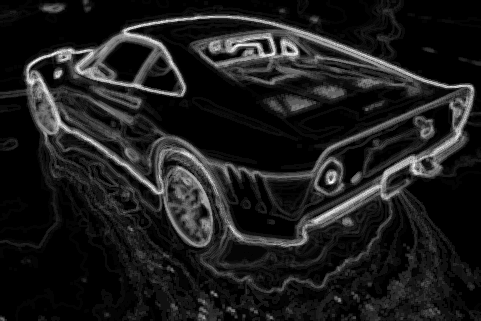} &
            \includegraphics[width = 0.15\textwidth]{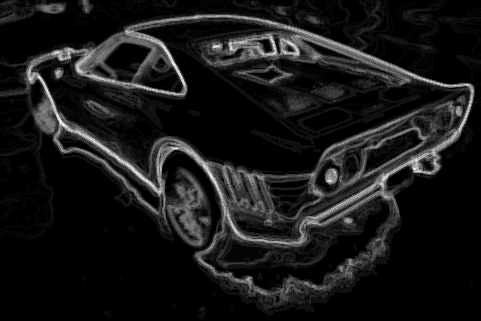} \\

            \includegraphics[width = 0.15\textwidth]{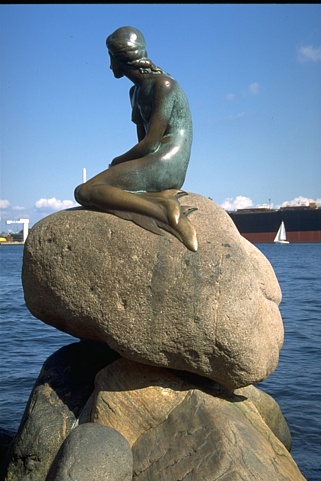} &
            \includegraphics[width = 0.15\textwidth]{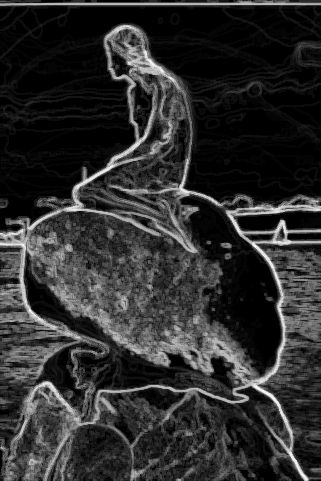} &
            \includegraphics[width = 0.15\textwidth]{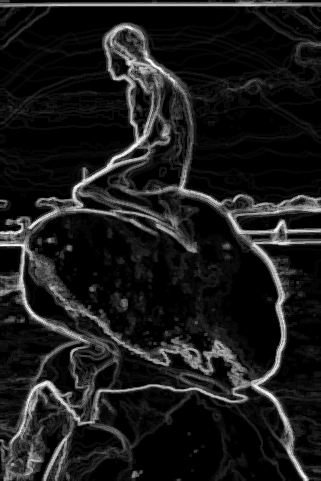} &
            \includegraphics[width = 0.15\textwidth]{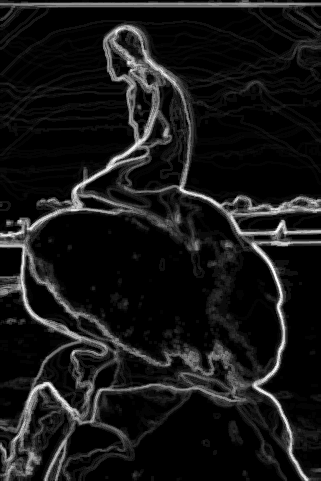} &
            \includegraphics[width = 0.15\textwidth]{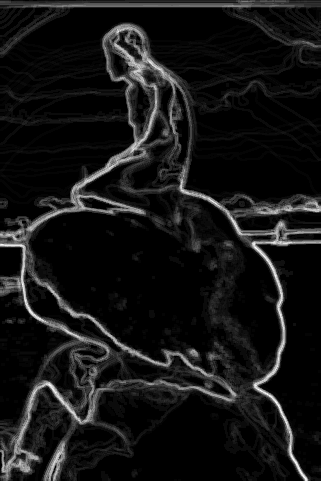} &
            \includegraphics[width = 0.15\textwidth]{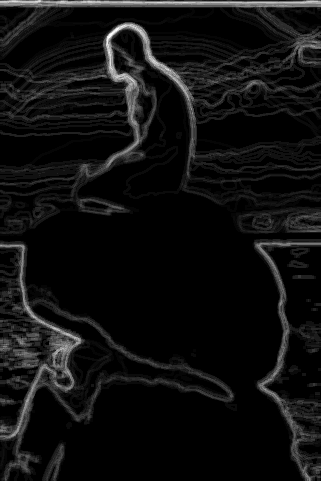} \\

            \includegraphics[width = 0.15\textwidth]{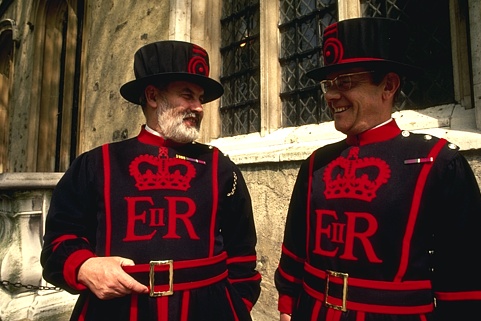} &
            \includegraphics[width = 0.15\textwidth]{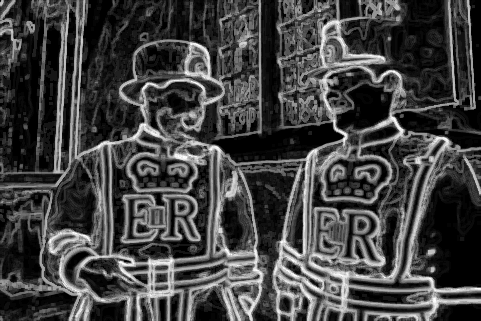} &
            \includegraphics[width = 0.15\textwidth]{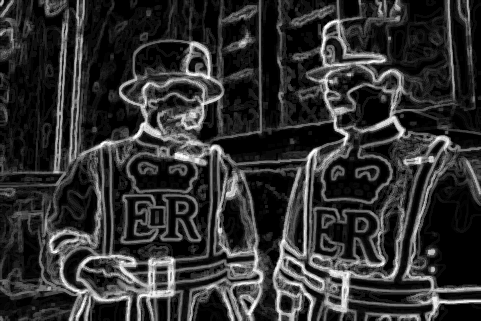} &
            \includegraphics[width = 0.15\textwidth]{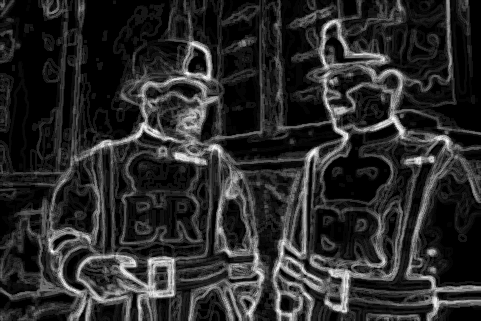} &
            \includegraphics[width = 0.15\textwidth]{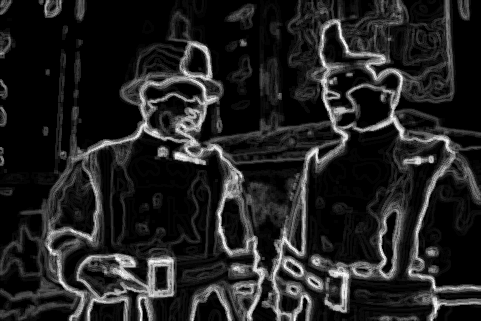} &
            \includegraphics[width = 0.15\textwidth]{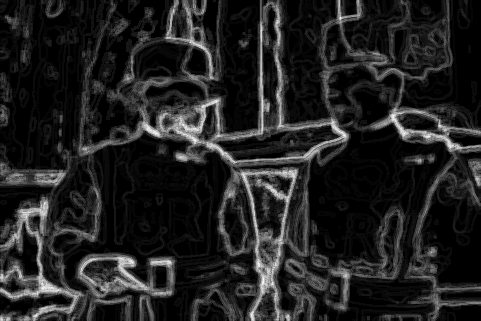} \\

            \includegraphics[width = 0.15\textwidth]{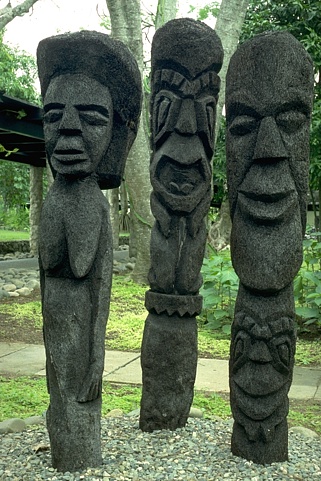} &
            \includegraphics[width = 0.15\textwidth]{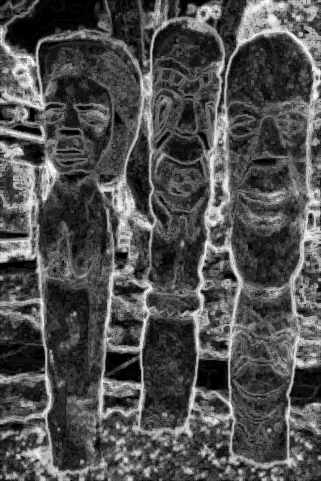} &
            \includegraphics[width = 0.15\textwidth]{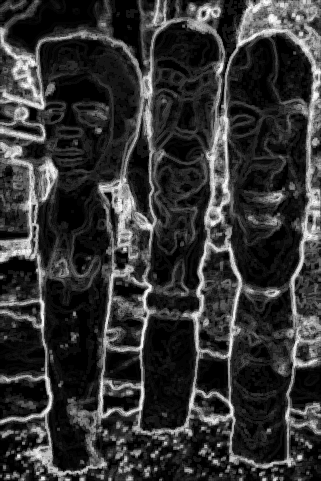} &
            \includegraphics[width = 0.15\textwidth]{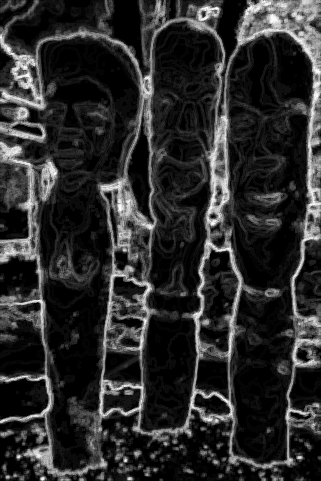} &
            \includegraphics[width = 0.15\textwidth]{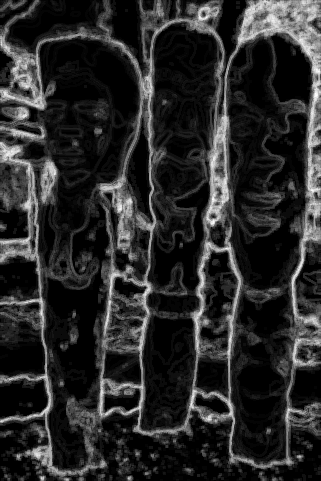} &
            \includegraphics[width = 0.15\textwidth]{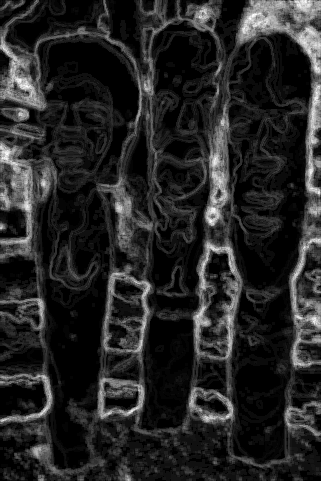} \\

            \includegraphics[width = 0.15\textwidth]{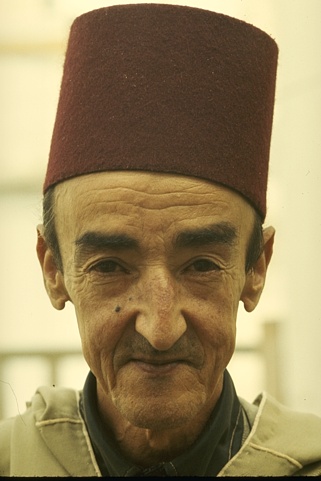} &
            \includegraphics[width = 0.15\textwidth]{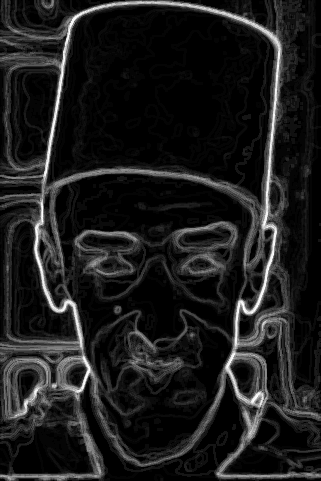} &
            \includegraphics[width = 0.15\textwidth]{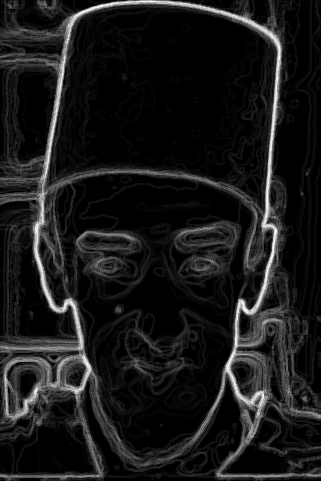} &
            \includegraphics[width = 0.15\textwidth]{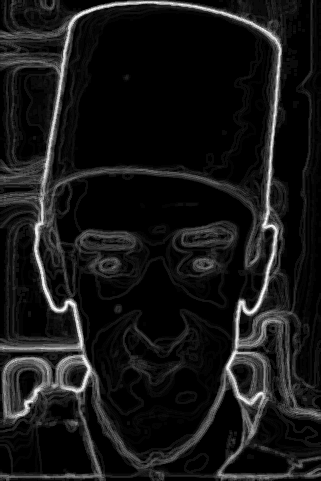} &
            \includegraphics[width = 0.15\textwidth]{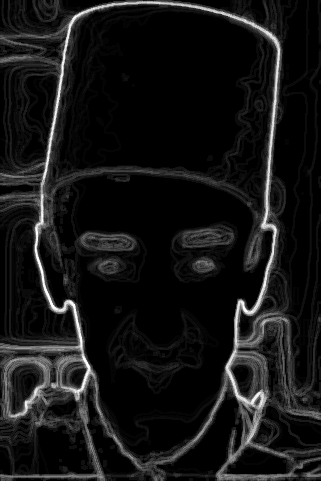} &
            \includegraphics[width = 0.15\textwidth]{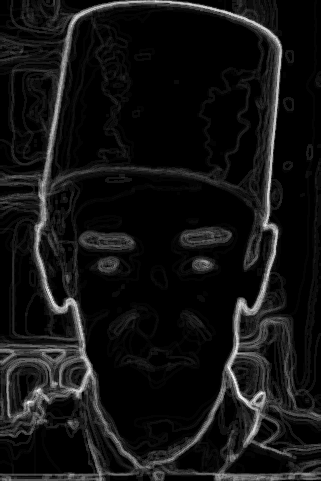} \\

            (a) Image & (b) $\lambda=0$ & (c) $\lambda=8$ & (d) $\lambda=16$ & (e) $\lambda=32$ & (f) $\lambda=\infty$ \\
            &

        \end{tabular}   \\ 
    \caption{\small \textbf{Contours for different $\lambda$ values}. (a) input image, (b) to (e) contours generated by our SCRF with $\lambda$ values ranging from $0$ to $\infty$. It can be seen, as expected, that as $\lambda$ grows larger the contours represent lager and smoother segments.}
    \label{fig_seg_example_1}
\end{figure*}

To illustrate the effect of the spatial coherency term we show, in figure \ref{fig_seg_example_1}, the contours generated using SCRF on a few images for a range of $\lambda$ values. This example clearly shows how larger values of $\lambda$ result in smoother segmentations and higher level contours.

We evaluated SCRF on the Berkley BSD300 and BSD500 datasets. All parameters were fixed throughout the evaluation. Contour accuracy was tested using the algorithm provided with the dataset. The input to our algorithm consisted of a 6 layered image, the first 3 layers were the original image converted to the Lab color space. Layers 4 to 6 were the same Lab image  filtered with a bilateral filter with a spatial standard deviation of 3 pixels and a intensity standard deviation equal to $10\%$ of the maximal image intensity range.

Our results on the BSD300 and BSD500 datasets can be seen in table \ref{tab_BDS_res}. The results indicate that our choice of $\lambda$ has improved the quality of the contours generated by our SCRF compared to the results of a regular Random Forest by about $10\%$ when choosing the optimal scale for the entire dataset (ODS). In addition our results are comparable to algorithms such as \cite{cour2005spectral}, \cite{felzenszwalb2004efficient} and \cite{sharon2006hierarchy}.

\begin{table} \label{tab_BDS_res}
\centering
\resizebox{0.9\textwidth}{!}{\begin{minipage}{\textwidth}
\begin{tabular}{|l|c|c|c||c|c|c|}
\cline{2-7}
\multicolumn{1}{c|}{} & \multicolumn{3}{|c||}{\textbf{BSD300}} & \multicolumn{3}{|c|}{\textbf{BSD500}}\\
\cline{2-7}
\multicolumn{1}{c|}{}           & ODS   & OIS  & AP   & ODS   & OIS   & AP   \\
\hline
\cite{arbelaez2011contour} & 0.67  & 0.71 &  &    &   &  \\
\hline
\cite{andres2011probabilistic} & 0.67  & 0.70 &  &   &   &  \\
\hline
\cite{cour2005spectral}         & 0.62  & 0.66 & 0.43 & 0.64  & 0.68  & 0.45 \\
\hline
\cite{felzenszwalb2004efficient}& 0.58  & 0.62 & 0.53 & 0.61  & 0.64  & 0.53 \\
\hline
SCRF $\lambda\sim N(8,4)$            & 0.57  & 0.62 & 0.51 & 0.6   & 0.64  & 0.53 \\
\hline
SCRF $\lambda=8$                     & 0.57  & 0.62 & 0.51 & 0.6   & 0.63  & 0.52 \\
\hline
\cite{sharon2006hierarchy}      & 0.56  & 0.59 & 0.54 &   &   &  \\
\hline
RF                     & 0.53  & 0.61 & 0.4  & 0.55  & 0.62  & 0.42 \\
\hline
\end{tabular}
\end{minipage}}
\smallskip
\caption{BSD300 and BSD500 Results. Shown are the F-measures when choosing an optimal scale for the entire dataset (ODS) or per image (OIS), as well as the average precision (AP). $\lambda=0$ corresponds to a segmentation forest without spatial coherence. $\lambda=8$ is a spatially coherent segmentation forest. }
\end{table}

\begin{figure} [t]
    \centering 
        \begin{tabular}{c c }

            \includegraphics[width = 0.23\textwidth]{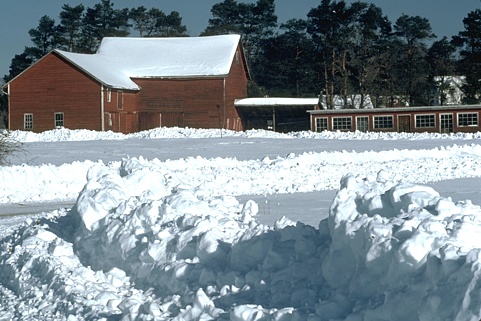} &
            \includegraphics[width = 0.23\textwidth]{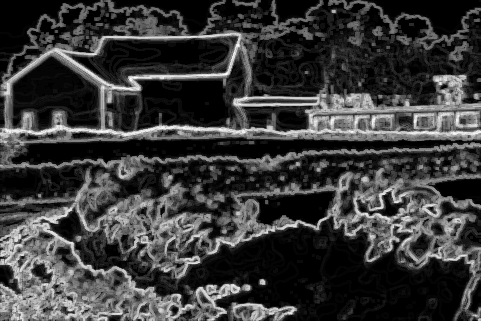} \\

            \includegraphics[width = 0.23\textwidth]{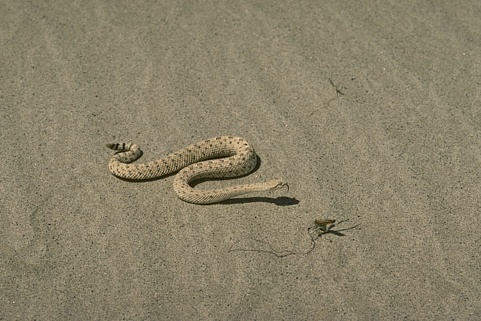} &
            \includegraphics[width = 0.23\textwidth]{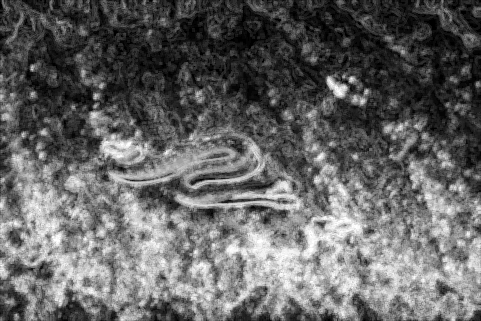} \\

            \includegraphics[width = 0.23\textwidth]{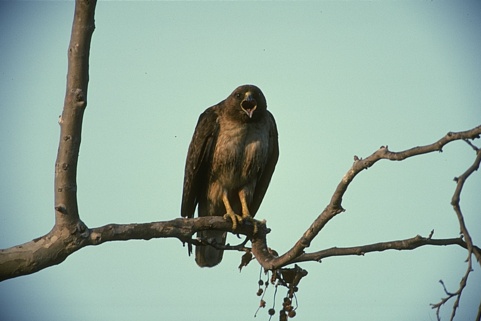} &
            \includegraphics[width = 0.23\textwidth]{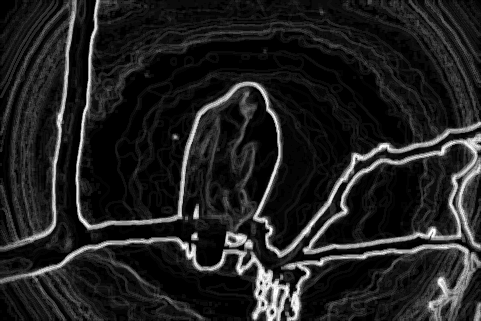} \\

            \includegraphics[width = 0.23\textwidth]{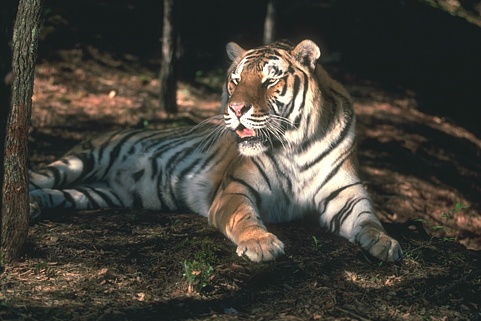} &
            \includegraphics[width = 0.23\textwidth]{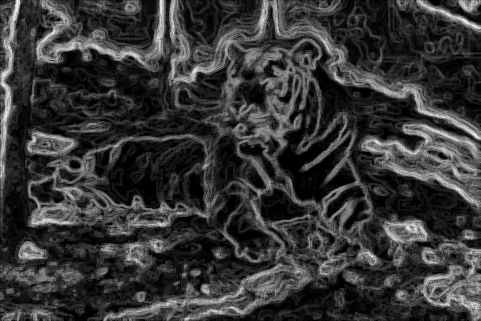} \\

                        (a) Image & (b) Contour\\

        \end{tabular}   \\ 
    \caption{\small \textbf{Successful and unsuccessful contour detection}. (a) input image, (b) contour generated using our SCRF. In the first and second rows are presented the images with the highest and lowest precisions respectively, the third and forth rows present the images with the highest and lowest recalls respectively.  }
    \label{fig_seg_example_3}
\end{figure}

\begin{figure} [t]
    \centering 
        \begin{tabular}{c}

             \includegraphics[width = 0.45\textwidth]{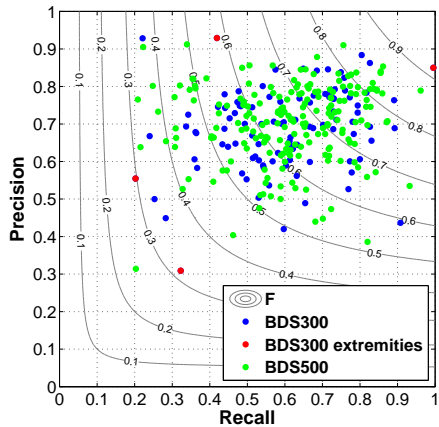} \\

        \end{tabular}   \\ 
    \caption{\small \textbf{BSD300 and BSD500 scatter}. Presented are the all test images in both datasets. The images marked in red are the images that got the highest and lowest recall and precision in the BSD300 test set, these images and their contours are shown in figure \ref{fig_seg_example_3} }
    \label{fig_seg_example_4}
\end{figure}

Figure \ref{fig_seg_example_3} shows the BDS300 images with highest and lowest precision and recall scores. SCRF assumes that the color of a segment can be described well with a GMM. As a result, it works well on rows 1 and 3. The algorithm struggles with textured regions. In particular, row 4 shows an example where the background has a very strong texture with high contrast. Figure \ref{fig_seg_example_4} shows a scatter plot of the precision and recall of all images in the BSD300 and BSD500 with the extremities of the BSD300 marked in red.

Our implementation uses non-optimized MATLAB code where each tree is constructed in parallel in a different thread. Constructing a forest took about 240 seconds per image on a Xeon E6550 2GHz CPU. Both tree construction and split function evaluation can be done in parallel on a GPU.

The spatial coherency term can be evaluated at a very low computational cost using the integral image. For each split function two integral images are evaluated, one for the pixels assigned to the left child and one for the pixels assigned to the right child of a given node. The spatial coherency term can then be evaluated for each pixel with O(1) complexity using these integral images.

\section{Conclusions}

We extended Random Forest to support spatial reasoning. The new forest, termed Spatially Coherent Random Forest (SCRF) encourages spatially coherent labeling. Each split function in SCRF is evaluated based on a traditional information gain measure that is regularized by a spatial coherency term. We have used SCRF for contour detection in natural images which is a surrogate to image segmentation. Each tree in the forest produces a segmentation of the image plane and the boundaries of the segmentations of all trees are aggregated to produce a final hierarchical contour map. The proposed modification is easy to implement and does not hurt the good properties of Random Forests that made them so popular.
We evaluated SCRF on the standard Berkeley Segmentation Datasets and found that they improved performance of regular Random Forests by \mbox{about $10\%$.}

{\small
\bibliographystyle{ieee}
\bibliography{SCRF_bib}
}

\end{document}